%% file: main.tex
\pdfoutput=1
\PassOptionsToPackage{table}{xcolor}
\documentclass{article}

\usepackage{iclr2027_conference,times}
\usepackage[normalem]{ulem}

\usepackage[utf8]{inputenc}
\usepackage[T1]{fontenc}
\usepackage{hyperref}
\hypersetup{colorlinks=true, linkcolor=blue!60!black, citecolor=blue!60!black, urlcolor=blue!60!black}
\usepackage{url}
\usepackage{booktabs}
\usepackage{amsfonts}
\usepackage{microtype}
\usepackage[table]{xcolor}
\usepackage{graphicx}
\usepackage{adjustbox}
\usepackage{amsmath}
\usepackage{amssymb}
\usepackage{mathtools}
\usepackage{amsthm}
\usepackage{tabularx}
\usepackage[capitalize,noabbrev]{cleveref}
\usepackage{ragged2e}
\usepackage{enumitem}
\usepackage{subcaption}
\usepackage[most]{tcolorbox}
\usepackage{placeins}
\usepackage{pifont}
\usepackage{float}
\usepackage{needspace} 
\tcbset{vctpanel/.style={
  colback=white,
  colframe=black!55,
  colbacktitle=black!55,
  coltitle=white,
  boxrule=0.4pt,
  arc=1mm,
  left=1.2mm,right=1.2mm,
  top=1.2mm,bottom=1.2mm,
  fonttitle=\bfseries\small,
  fontupper=\footnotesize,
  before skip=4pt,after skip=4pt
}}

\graphicspath{{images/}}
\theoremstyle{plain}

\theoremstyle{definition}

\theoremstyle{remark}

\newcommand{\cmark}{\ding{51}}
\newcommand{\xmark}{\ding{55}}

\title{Distilling Visual Reasoning into Text Space}
\author{Wenhan Yang$^{1}$, Nilay Naharas$^{1}$, Ali Payani$^{2}$, Baharan Mirzasoleiman$^{1}$ \\
$^{1}$University of California, Los Angeles \quad $^{2}$Cisco Research \\
\texttt{hangeryang18@g.ucla.edu}}

\iclrfinalcopy

\begin{document}
\maketitle

\begin{abstract}
\input{abstract}
\end{abstract}

\input{introduction}
\input{related}
\input{preliminary}
\input{method}
\input{experiment}
\input{ablation}
\input{analysis}
\input{conclusion}

\subsection*{AI use statement}
We used generative AI tools solely to improve the manuscript’s writing, including grammar, phrasing, and clarity. All AI-assisted text was carefully reviewed and edited by the authors. Generative AI was not used to develop research ideas, formulate theoretical claims, or produce experimental results.

\bibliography{references}
\bibliographystyle{iclr2027_conference}
\newpage 
\appendix
\input{appendix}
\end{document}

%% file: abstract.tex
Large Vision–Language Models (LVLMs) have shown strong promise for multimodal reasoning, yet often struggle with tasks requiring concepts beyond what is directly observable in the input image. Existing methods generate intermediate images or latent visual tokens to guide reasoning, but these representations can introduce errors and increasingly interfere with textual reasoning as reasoning progresses. We propose Visual-to-Text Chain-of-Thought Distillation (V2T), a framework that enables LVLMs to internalize visual reasoning \textit{without} generating intermediate visual representations at inference time. V2T first trains a teacher LVLM using interleaved visual and textual chains of thought, and then uses knowledge distillation to train a student LVLM using the teacher’s logits and cross-entropy supervision from ground-truth textual reasoning. When reasoning images can be mapped to the original image, V2T can additionally distill the teacher’s attention to corresponding regions, while ground-truth bounding boxes can further guide a subsequent reinforcement learning stage. Experiments across multiple multimodal reasoning benchmarks show that V2T consistently outperforms the teacher and existing baselines, improving average accuracy by 14.3\% on a held-out set and 2.7\% on the broader visual evaluation suite. Moreover, lightweight SFT and substantially reduced RL make V2T up to 42$\times$ faster to train than state-of-the-art baselines.
\looseness=-1

%% file: introduction.tex
\section{Introduction}
Large vision–language models (LVLMs) have demonstrated strong performance on multimodal reasoning tasks involving an input image and textual question by generating textual chains of thought (CoTs). This is often achieved through supervised fine-tuning on long textual CoTs conditioned on the input image, followed by reinforcement learning to encourage step-by-step reasoning \citep{shao2024visual, xu2025llava, zhang2025r1,huang2025vision,wang2025vl,zhang2025improve,chen2025sft}.
However, many problems cannot be solved from the raw image alone. For instance, as shown in Fig.~\ref{fig:zebra_cot_interleaved_examples}, solving geometric problems often requires auxiliary constructions, such as additional lines or points, to reveal implicit relationships (e.g., perpendicular distances or tangency) that are not apparent in the original image. Humans naturally imagine such intermediate visual states, but LVLMs trained primarily in the text space often struggle with reasoning over such representations \citep{li2025latent,li2025imagine,xu2025visual}.\looseness=-1

\par\vspace{3pt}\noindent
\begin{minipage}{\linewidth}
\centering
\captionsetup{type=figure,font=small,skip=4pt}
\includegraphics[width=\linewidth]{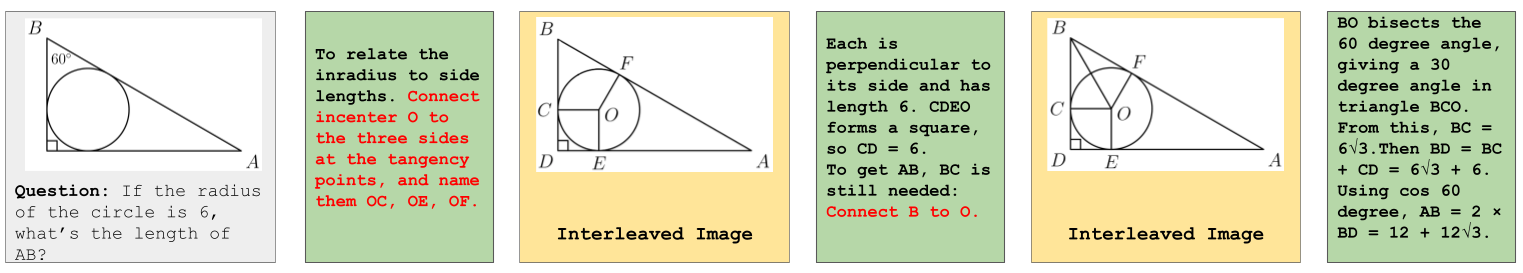}
\caption{Example of multimodal CoT, where every step of thinking is paired with an image.}
\label{fig:zebra_cot_interleaved_examples}
\end{minipage}
\par\vspace{4pt}

To provide such intermediate visual information, recent work interleaves visual information directly into the reasoning process. One line augments LVLMs with external visual tools (e.g., sketchpads or pixel-level operations), allowing models to manipulate images during reasoning via actions such as zooming, drawing auxiliary lines, or highlighting relevant regions \citep{hu2024visual, qi2024cogcom, hu2024visualdistill, wu2025vtool, su2025openthinkimg, huang2025visualtoolagent, wu2023visual, yang2023mm}. Another line focuses on latent visual reasoning, training models on interleaved multimodal data where textual CoTs are paired with reasoning images, and learning visual tokens that act as surrogates for intermediate images at inference time \citep{su2025pixel, wang2025monet, li2025latent, tong2025sketch, dong2025interleaved, yang2025machine}.

\begin{figure}[t]
    \centering
    \includegraphics[width=\linewidth, height=0.22\linewidth]{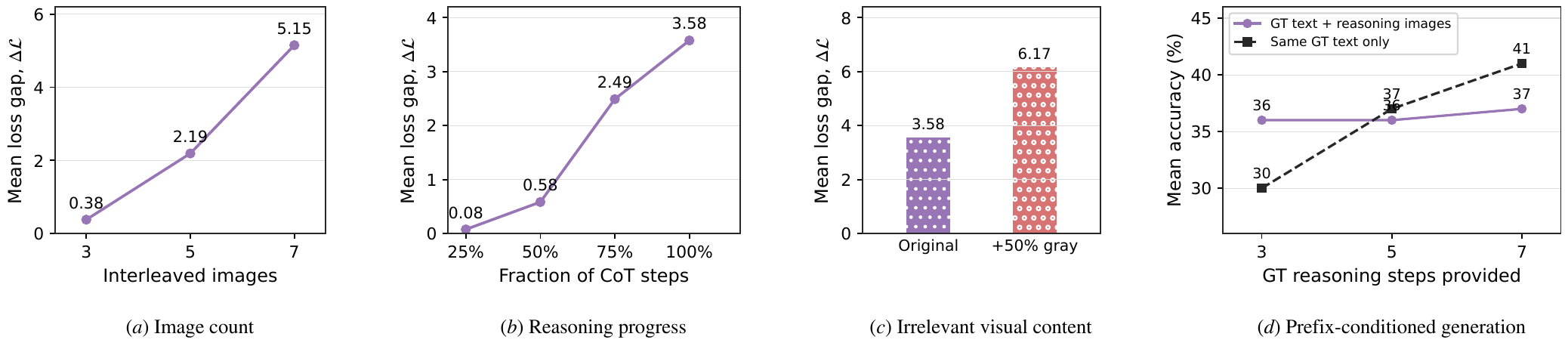}
    \vspace{-5mm}
    \caption{
    \textbf{Interleaved visual inputs increasingly interfere with textual reasoning.}
(a–c) The loss gap between reasoning with and without interleaved images widens as reasoning progresses and further increases with irrelevant visual content.
(d) With longer reasoning prefixes, text-only continuation matches or outperforms continuation with interleaved images.
\looseness=-1
    }
    \label{fig}
    \vspace{-4mm}
\end{figure}

Despite their promise, these approaches face two key limitations. First, generated intermediate images or latent visual tokens can be inaccurate or incomplete, potentially misleading subsequent reasoning and propagating errors \citep{li2025look, wang2025illusion,liu2025faithfulness}. Moreover, as shown in Fig. \ref{fig}, interleaved visual inputs can increasingly interfere with textual reasoning as the reasoning process unfolds: the loss gap between reasoning with and without interleaved images widens over successive steps, and even adding irrelevant visual content, such as a gray background, further increases this gap (see Appendix~\ref{app:motivation_analysis} for details). This suggests that interleaved visual representations can progressively disrupt the model’s textual reasoning, even when they provide little additional useful information.

These observations motivate a central question: \textit{can visual reasoning be internalized into text-based CoT, so that models can reason as if they had access to intermediate visual information without actually generating it?} We propose {Visual-to-Text Chain-of-Thought Distillation (V2T)}, which first trains a teacher on multimodal reasoning trajectories with intermediate reasoning images, then distills its reasoning into a student that reasons entirely in text space. The student is trained using the teacher's logits and ground-truth textual CoT, with optional attention distillation when reasoning images can be mapped to the original image, e.g. crop or transformation. When ground-truth regions are available, a short subsequent reinforcement learning stage, e.g., GRPO, can further encourage attention to task-relevant regions. At inference, the student reasons without intermediate images, avoiding their interference and error propagation while retaining the teacher's visual reasoning capabilities. V2T requires lightweight SFT for interleaved teacher training and student distillation, while the strong initialization from distillation substantially reduces the amount of subsequent RL, resulting in up to 42$\times$ faster training than baselines, even with GRPO. Remarkably, the student consistently outperforms its teacher.

Through extensive evaluation on in-domain and broad multimodal reasoning benchmarks, we show that V2T consistently outperforms both external visual-tool and latent visual-reasoning methods. On a held-out set, V2T achieves 75.5\% average accuracy, improving by 14.3\% over its teacher and 14.6\% over the strongest existing baseline. On general benchmarks, V2T achieves an overall score of 61.8\%, outperforming the strongest baseline by 2.7\%. Attention and error analyses further show that V2T attends more strongly to task-relevant regions and makes fewer reasoning errors than both text-only CoT and the teacher, indicating stronger visual grounding and more reliable reasoning without intermediate images. Notably, V2T's largest gains occur in spatial and global-context reasoning ($-16.1\%$), highlighting the benefits of internalized visual grounding for scene understanding.\looseness=-1

%% file: related.tex
\section{Related Work}

Existing visual CoT approaches %
can be broadly grouped into the following three categories.

\textbf{Text-Only Visual Chain of Thoughts}
The first category focuses on constructing datasets where images are paired with detailed textual reasoning steps \citep{qiao2024prism, shao2024visual,xu2025llava,hu2022promptcap}. During training, the model learns to observe the image, verbalize its perception, and reason about it before arriving at a final decision. 
However, a key limitation is that reasoning can become dominated by long text sequences, overshadowing the underlying visual information. \looseness=-1

\textbf{Tool-Augmented Visual Chain of Thoughts}
The second category extends beyond textual reasoning by incorporating visual tools, such as drawing bounding boxes, sketching shapes, cropping, or zooming into salient regions \citep{hu2024visual,qi2024cogcom,hu2024visualdistill,wu2025vtool,su2025openthinkimg, huang2025visualtoolagent, wu2023visual,yang2023mm}. These tools generate intermediate images or annotations that are fed back into the model for further reasoning. While this can produce more precise and interpretable reasoning, it often requires strong models or reinforcement learning to reliably learn tool-use policies, limiting scalability. Moreover, reasoning is constrained by tool capabilities, and inaccuracies in intermediate images can propagate through the autoregressive process and degrade performance.

\textbf{Interleaved Latent Visual Chain of Thoughts}
A third line of work integrates visual information directly into the model’s reasoning process. MINT-CoT \citep{chen2025mint} learns interleaved tokens corresponding to specific image regions, grounding each reasoning step in localized visual evidence. PixelReasoner \citep{su2025pixel} maintains a latent visual state that is iteratively updated to refine fine-grained details. Other approaches \citep{li2025latent, wang2025monet} predict latent visual representations that can be decoded into approximate image regions, helping the model refocus on relevant areas. Finally, unified latent-space methods reason directly within shared visual representations, enabling manipulation of abstract visual features beyond the original input.
However, these interleaved visual inputs can increasingly interfere with textual reasoning as
the reasoning process unfolds, harming the performance (cf.\ Fig.~\ref{fig}).

We propose V2T, which distills visual reasoning from a teacher trained with interleaved images into a student that reasons entirely in text space. By avoiding intermediate visual representations during inference, V2T retains visual reasoning capabilities without the modality switching, image-generation errors, and error propagation of existing approaches. In doing so, it achieves superior performance.

%% file: preliminary.tex
\begin{figure*}[t!]
    \centering
    \vspace{-4mm}
    \includegraphics[width=\linewidth, height=0.43\linewidth]{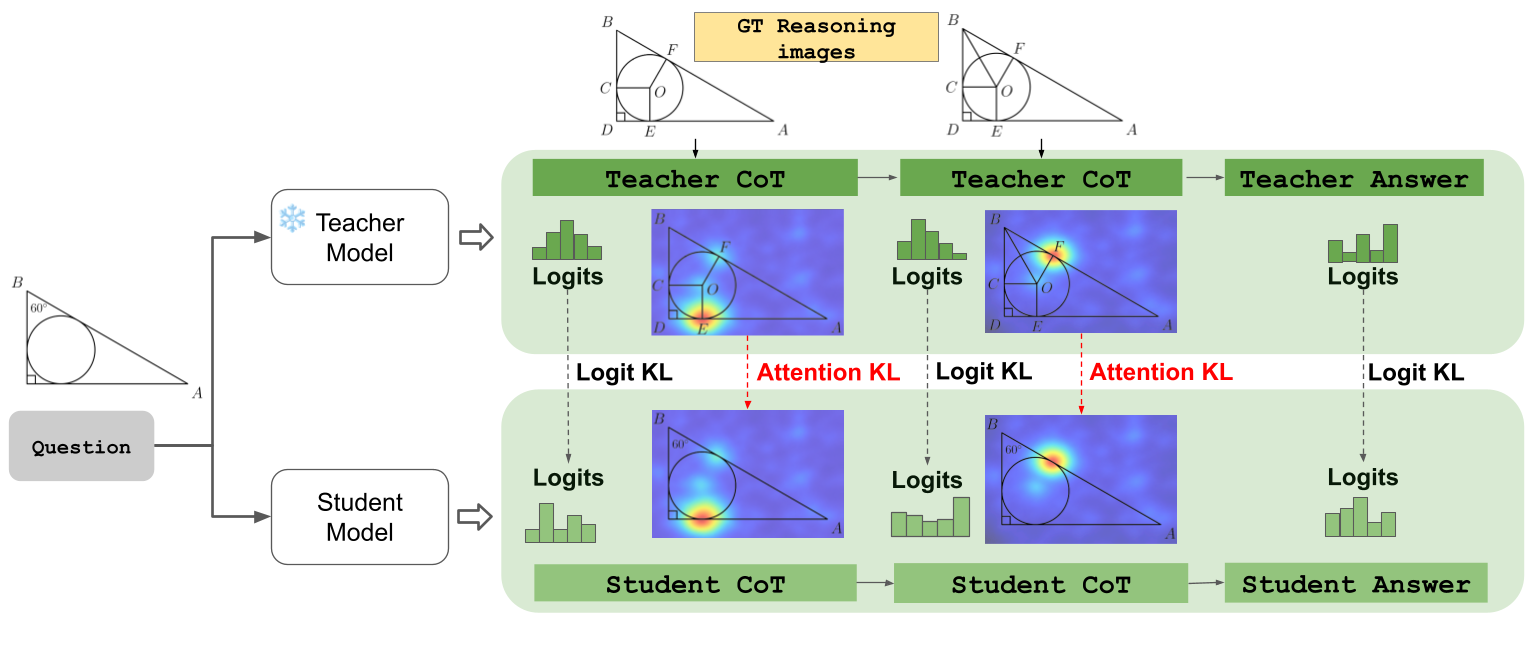}
    \vspace{-8mm}
    \caption{\textbf{V2T Training.} V2T first trains a teacher with interleaved images to ground its CoT in visual cues. The student model, which does not have access to interleaved images, is trained using KL divergence to match the teacher’s textual CoT and logits, along with attention distillation to transfer the teacher’s visual grounding, and cross-entropy loss to match the ground-truth textual CoT.}
    \label{fig:attn_coincide_map}
    \vspace{-2mm}
\end{figure*} 
\section{Distilling Visual Chains of Thought into Text Space}
\begin{figure*}[t] %
    \centering
    \includegraphics[width=\linewidth]{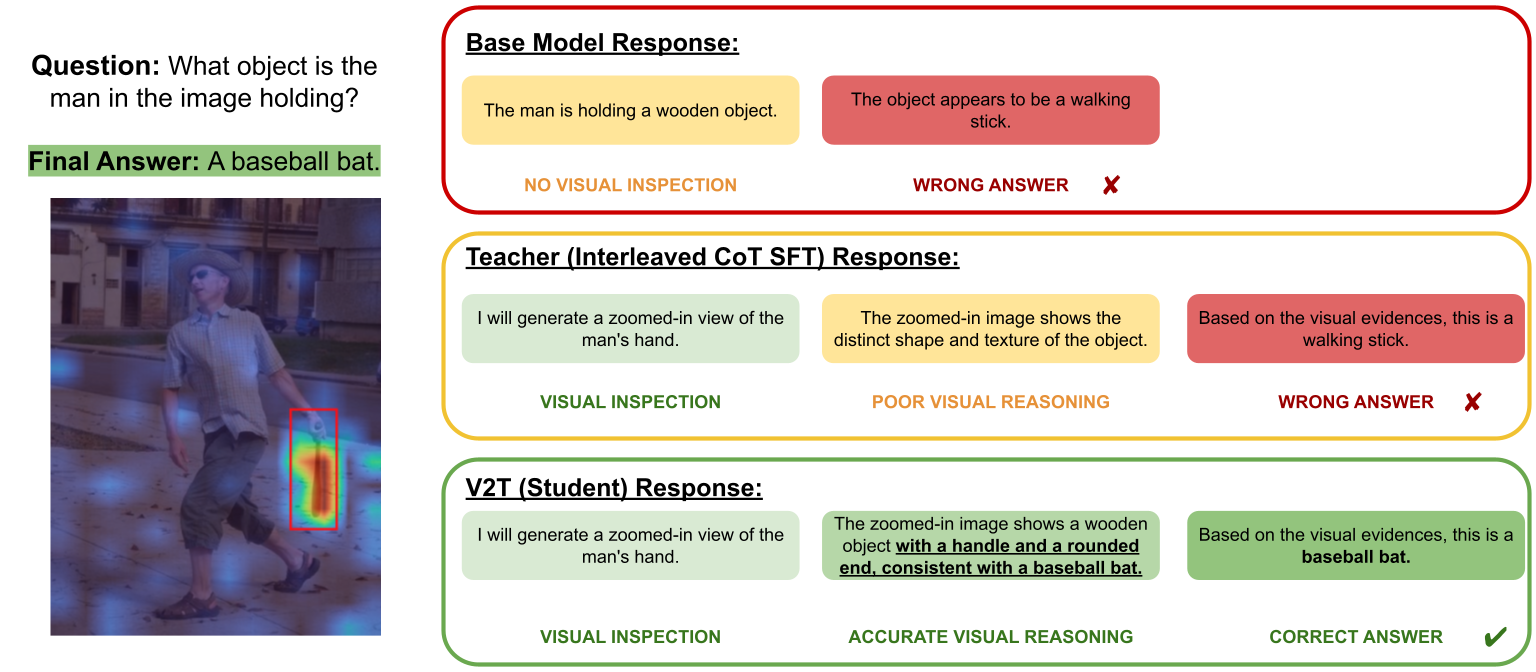}
    \caption{\textbf{V2T inference does not require intermediate images.} At inference, the base model, teacher (Interleaved CoT SFT), and V2T student (without attention distillation or GRPO) receive the question and original problem image and generate textual responses. The base model and teacher misidentify the object as a walking stick, while the student correctly identifies a baseball bat. Phrases such as ``zoomed-in view'' are inherited from the teacher's interleaved CoT format; no intermediate image is generated or provided, and V2T reasons only over the original image. %
    }

    \label{fig:reasoning_traces}
    \vspace{-2mm}
\end{figure*}

In this section, we introduce V2T, which enables LVLMs to reason over the original problem image \textit{as if} intermediate reasoning images were available, while generating a purely textual CoT at inference (Fig.~\ref{fig:attn_coincide_map}; an inference example is shown in Fig.~\ref{fig:reasoning_traces}). 

\subsection{Reasoning with Large Vision-Language Models}
An LVLM consists of a vision encoder and a large language model (LLM) that jointly process visual and textual inputs. 
Given a text prompt and an input image, the model first encodes the image and prompt into visual $v$ and text $x$ tokens, then performs autoregressive reasoning by modeling: \vspace{-2mm}
\begin{equation}
p_{\theta}(y_{1:T} \mid v, x)
= \prod_{t=1}^{T} p_{\theta}\!\left(y_t \mid v, x, y_{<t}\right),
\end{equation}
where $y_{<t}$ are intermediate text CoT tokens. %

%% file: method.tex
\textbf{Limitations of LVLMs Reasoning. }
As shown in Fig.~\ref{fig:zebra_cot_interleaved_examples} and Fig.~\ref{fig:zebra_case_example2} in the Appendix,
many challenging tasks %
benefit from pairing each CoT step with a corresponding \textit{reasoning image}. %
However, standard VLMs operate autoregressively over discrete tokens and lack a reliable mechanism to generate faithful intermediate visual states. %
As illustrated in Fig. \ref{fig}, intermediate images or visual tokens can increasingly interfere with textual reasoning as reasoning progresses. Without strong supervision, errors can propagate through autoregressive reasoning and degrade performance.

\subsection{V2T: Distilling Visual Chains of Thought into Text Space} 
To address these limitations, our key idea is to enable LVLMs to reason in text space, \textit{as if} grounded in rich, step-by-step visual information. V2T consists of two stages. In Stage 1, it trains a teacher on CoT steps paired with intermediate reasoning images, grounding its reasoning in visual information. In Stage 2, it uses {knowledge distillation} to transfer this visual reasoning to a student that reasons entirely in text space, using the teacher’s logits and ground-truth textual CoT. When reasoning images can be mapped to the problem image, it additionally distills the teacher’s attention to corresponding regions, followed by RL when ground-truth regions are available. V2T retains visual grounding while avoiding intermediate representations and error propagation at inference.
Next, we describe V2T in detail.\looseness=-1

\subsubsection{Stage 1: Training the Teacher Model on Multimodal Reasoning Data}
Let $\mathcal{D}$ denote a dataset of interleaved multimodal reasoning examples. Each example consists of a text prompt and an initial problem image, with corresponding textual and visual tokens $x$ and $v_{\text{prob}}$. It further contains a $K$-step CoT, where each step $t\in[K]$ consists of ground-truth text $y_t$ and may be paired with an intermediate reasoning image with visual tokens $v_{\text{reason},t}$. The complete target text sequence is $y=(y_1,\ldots,y_K)$, including the textual CoT and final answer.

We train the teacher model $f_\theta$ using cross-entropy on textual tokens, conditioned on the preceding text-image pairs:
\begin{equation}
\label{eq:teacher-ce}
\mathcal{L}_{\mathrm{CE}}(\theta)
=-\sum_{t=1}^{K}
\log p_\theta
\left(
y_t
\mid
x, v_{\mathrm{prob}},
{(y_j, v_{\mathrm{reason},j})}_{j<t},
v_{\mathrm{reason},t}
\right).
\end{equation}
Here, $y_t$ denotes the ground-truth text at step $t$. Each reasoning image $v_{\mathrm{reason},t}$ serves as a conditioning input and receives no prediction loss. We use step-level notation for simplicity; in practice, the loss is computed token by token under teacher forcing.

\textbf{{Teacher limitation.}} While interleaved reasoning images provide critical information during training, they are unavailable at inference, creating a train–test mismatch that limits the teacher’s performance.
\looseness=-1

\subsubsection{Stage 2: Distilling Multimodal Reasoning into the Text Space}
To address this limitation, in the second stage we train a student LVLM that reasons entirely in text space, without access to intermediate images, while retaining the visual reasoning learned by the teacher. We freeze the teacher $f_\theta$ and distill its next-token logits to train a student $f_\phi$ to generate textual reasoning from the prompt $x$ and original problem image $v_{\mathrm{prob}}$. The student receives no intermediate reasoning images during training or inference. The teacher additionally conditions on the current reasoning image and preceding interleaved text-image pairs. 
Both models use the same ground-truth textual prefix under teacher forcing:
\begin{align}
\label{eq:next-token}
q_\theta(y_t)
&=
p_{\theta}\left(
y_t \mid
x,v_{\mathrm{prob}},
v_{\mathrm{reason},t},
{(y_j,v_{\mathrm{reason},j})}_{j<t}
\right),
&\quad&\text{Teacher},\\
p_\phi(y_t)
&=
p_{\phi}\left(
y_t \mid
x,v_{\mathrm{prob}},
y_{<t}
\right),
&&\text{Student}.
\end{align}
Here, $y_t$ denotes the ground-truth text at step $t$, and $y_{<t}$ denotes the preceding textual CoT steps. 

The logits distillation loss is therefore:
\begin{table*}[t]
\centering
\caption{Zebra-CoT held-out set. Best and second-best results across all methods are highlighted in \textbf{bold} and \underline{underline}, respectively. As Bagel-Zebra is trained on the full Zebra-CoT dataset, we exclude it from the evaluation. V2T achieves the best overall performance of $75.5\%$, outperforming the interleaved teacher model by $14.3\%$.}
\label{tab:zebra_cot}
\small
\setlength{\tabcolsep}{6pt}
\begin{tabular}{lccccc}
\toprule
Method & Graph Algorithm & RPM & Tetris & Visual Search & Avg \\
\midrule

\multicolumn{6}{l}{\textit{LVLMs with text-only visual CoT}} \\[0.3ex]

R1-VL-7B
& 44.5 & 10.6 & 55.0 & 60.5 & 42.6 \\
VLAA-Thinker-7B
& \underline{58.0} & 54.8 & 68.0 & 63.0 & 60.9 \\
VL-Rethinker-7B
& 45.0 & 55.8 & \underline{69.0} & 64.5 & 58.6 \\

\addlinespace[0.8ex]
\midrule
\multicolumn{6}{l}{\textit{LVLMs with interleaved latent visual CoT}} \\[0.3ex]

LVR-7B
& 17.5 & 23.1 & 25.5 & \underline{66.5} & 33.2 \\
MONET-7B
& 39.0 & 7.5 & 27.5 & 59.0 & 33.3 \\

\addlinespace[0.8ex]
\midrule
\multicolumn{6}{l}{\textit{Interleaved SFT and baseline}} \\[0.3ex]

Qwen2.5-VL-7B
& 54.5 & 45.7 & \underline{69.0} & 61.0 & 57.6 \\
Interleaved SFT
& 48.5 & \underline{65.8} & 65.0 & 65.5 & \underline{61.2} \\
\rowcolor{gray!15}
\textbf{V2T (Ours)}
& \textbf{61.5} & \textbf{81.9} & \textbf{81.5} & \textbf{77.0} & \textbf{75.5} \\

\bottomrule
\end{tabular}
\vspace{-0.6em}
\end{table*}
\begin{equation}
\hspace{1.5cm}
\mathcal{L}_{\mathrm{KD}}(\phi) =
\sum_t
\mathrm{KL}\left(
q_\theta(y_t) \,\middle\|\, p_\phi(y_t)
\right).
\hspace{2cm}
\text{Teacher logit distillation}
\end{equation}
\textbf{\textit{Remark.}} During distillation, the teacher is conditioned on the ground-truth interleaved reasoning trajectory rather than generated intermediate images, preventing errors from generated visual states from propagating through the distillation process.

We also supervise the student on the ground-truth textual tokens: \begin{equation} \label{eq:text} \hspace{1.5cm}\mathcal L_{\mathrm{text}}(\phi) =
-\sum_t
\log p_\phi\left(
y_t\mid y_{<t},x,v_{\mathrm{prob}}
\right).
\qquad
\text{Ground-truth supervision}
\end{equation}
The final V2T objective is as follows:
\begin{equation} \label{eq:total} \mathcal L_{\mathrm{total}}(\phi) =\mathcal L_{\mathrm{text}}(\phi) +\lambda\mathcal L_{\mathrm{KD}}(\phi). \end{equation}
where $\lambda$ controls the strength of the distillation signal. We use $\lambda=0.1$ in our experiments and ablate its effect in Appendix~\ref{app:hparam-ablation}.

\subsubsection{Auxiliary: Distilling Visual Grounding Via Attention}
\label{sec:attention_distillation}

When each intermediate reasoning image can be spatially mapped to a region of the original problem image, such as a crop or transformation, V2T can further distill the teacher’s attention as an auxiliary signal. For each reasoning step, we extract attention at the first ground-truth text token following the reasoning image, following prior work on probing visual attention at reasoning-step boundaries and the first generated token \citep{gao2025interleaved, chen2025spatial, li2026visual}. At this position, the reasoning image is fully available without intervening text. We project the teacher’s attention onto the corresponding region of the original image and align it with the student’s attention using a KL loss. This encourages attention to task-relevant regions and improves visual grounding.

Formally, let $\mathbf a_{\theta,t}$ and $\mathbf a_{\phi,t}$ denote the teacher's and student's normalized attention weight distribution to the reasoning image and problem image, respectively, at reasoning step $t$. Let $S \subseteq [K]$ be the reasoning steps for which $v_{\mathrm{reason},t}$ can be spatially mapped to a region of $v_{\mathrm{prob}}$.

We minimize their forward KL divergence:
\begin{equation}
\mathcal L_{\mathrm{attn}}(\phi)
=
\frac{1}{|S|}
\sum_{t\in S}
\mathrm{KL}\left(
\mathbf{a}_{\theta,t}
\,\middle\|\,
\mathbf{a}_{\phi,t}
\right).
\end{equation}
The V2T training objective becomes:
\begin{equation}
\mathcal L_{\mathrm{total}}^{\mathrm{attn}}(\phi)
=
\mathcal L_{\mathrm{text}}(\phi)
+
\lambda\mathcal L_{\mathrm{KD}}(\phi)
+
\beta\mathcal L_{\mathrm{attn}}(\phi),
\end{equation}
where $\beta$ controls the strength of attention supervision.

\subsubsection{Auxiliary: Reinforcement Visual Grounding}
\label{sec:attention_grounding_reward}

After Stage 2, with or without attention distillation, the V2T student learns to reason without intermediate images while retaining the teacher’s visual reasoning signals. Because distillation already provides a strong initialization with effective visual reasoning capabilities, only a short GRPO fine-tuning stage can be optionally applied to further improve the generalization performance. We use standard task rewards and, when ground-truth bounding boxes are available, an additional \emph{visual grounding reward} that encourages the student to attend to task-relevant image regions.

Specifically, for each reasoning step, we extract the student’s attention to the problem image to identify the attended region $\widehat{M}$ and compare it with the ground-truth region $M$. We use this alignment as a visual grounding reward:
\begin{equation}
r_{\mathrm{ground}}
=
\frac{1}{2}
\left(
\frac{|\widehat M \cap M|}{|\widehat M|}
+
\frac{|\widehat M \cap M|}{|M|}
\right).
\end{equation}
This reward balances how much of the attended region overlaps the target region with how much of the target region is covered. The final GRPO reward is:
\begin{equation}
r = r_{\mathrm{task}} + \eta r_{\mathrm{ground}},
\end{equation}
where $r_{\mathrm{task}}$ combines a binary answer-correctness reward and a format reward weighted by $0.1$. We set $\eta=0.5$; the effect of the visual grounding reward is evaluated in Sec.~\ref{sec:ablation}.

\begin{table*}[t]
    \centering
    \caption{Results on the broader visual evaluation suite. Best and second-best results are highlighted in \textbf{bold} and \underline{underline}, respectively. V2T achieves an average of 61.8, outperforming the strongest non-V2T baseline by 2.7 points; V2T + GRPO reaches 63.4. Training costs marked with $^{*}$ are estimated from the reported training configurations. V2T trains in 20 GPU-hours (80 with GRPO) at $1\times$ inference, compared with $2\times$ for latent reasoning, $30\times$ for Bagel-Zebra, and $45\times$ for tool-based methods.}
    \label{tab:zebra_table2}
    \small
    \setlength{\tabcolsep}{0.8pt}
    \renewcommand{\arraystretch}{1.08}
    \ifdefined\zebratw\else\def\zebratw{0pt}\fi
    \newsavebox{\zebratablebox}
    \sbox{\zebratablebox}{%
    \begin{tabular}{@{}lcccccc cccccc cc@{}}
    \toprule
    \textbf{Method}
    & \textbf{SFT}
    & \textbf{GRPO}
    & \textbf{Train.}
    & \textbf{HR4K}
    & \textbf{HR8K}
    & \textbf{V*}
    & \multicolumn{6}{c}{\textbf{BLINK}}
    & \textbf{POPE}
    & \textbf{Avg} \\
    \cmidrule(lr){8-13}
    &
    &
    & \textbf{(GPU-h)}
    &
    &
    &
    &
    Func. & Visual & Obj. Loc. & Foren. & Jig. & Multi.
    &
    & \\
    \midrule
    
    \multicolumn{15}{@{}c@{}}{\makebox[\zebratw][s]{\textit{LVLMs with text-only visual reasoning}\hfill\textit{Inference: $1\times$}}} \\[0.3ex]
    
    R1-VL-7B
    & \cmark & \cmark & $167^{*}$
    & 52.9 & 45.2 & 43.5
    & 22.3 & 52.3 & 48.4 & 44.7 & 65.3 & 42.1
    & 90.1 & 50.6 \\        
    
    VLAA-Thinker-7B
    & \xmark & \cmark & $167^{*}$
    & 50.5 & 41.4 & 48.7
    & 22.3 & 50.0 & 48.4 & 47.7 & 60.0 & 42.1
    & 86.1 & 49.7 \\
    
    VL-Rethinker-7B
    & \cmark & \cmark & 128
    & 61.0 & 48.9 & 64.4
    & 23.9 & 55.2 & 46.7 & 48.5 & 56.0 & 47.4
    & 86.4 & 53.8 \\

    Vision-R1-7B
    & \cmark & \cmark & 3392
    & \textbf{75.3} & \underline{70.5} & 80.6
    & 29.2 & 50.6 & 48.4 & \underline{50.0} & 59.3 & 39.1
    & 88.2 & 59.1 \\
    
    \addlinespace[0.8ex]
    \midrule
    
    \multicolumn{15}{@{}c@{}}{\makebox[\zebratw][s]{\textit{LVLMs with latent visual reasoning}\hfill\textit{Inference: $2\times$}}} \\[0.3ex]
    
    LVR-7B
    & \cmark & \cmark & 160
    & 69.4 & 63.9 & 77.5
    & 27.7 & 49.4 & \underline{53.3} & 40.2 & 50.0 & 50.4
    & 85.2 & 56.7 \\
    
    MONET-7B
    & \cmark & \cmark & $330^{*}$
    & 65.8 & 55.5 & 76.4
    & 26.2 & 55.8 & 50.8 & 36.4 & 55.3 & 50.4
    & 84.7 & 55.7 \\
    
    \addlinespace[0.8ex]
    \midrule
    
    \multicolumn{15}{@{}c@{}}{\makebox[\zebratw][s]{\textit{Unified interleaved model}\hfill\textit{Inference: $30\times$}}} \\[0.3ex]
    
    Bagel-Zebra-7B
    & \cmark & \xmark & $110^{*}$
    & 62.3 & 51.4 & 64.4
    & \underline{33.9}
    & \textbf{70.9}
    & \textbf{60.7}
    & 26.5
    & \textbf{74.0}
    & 46.6
    & 87.0 & 57.8 \\
    
    \addlinespace[0.8ex]
    \midrule
    \multicolumn{15}{@{}c@{}}{\makebox[\zebratw][s]{\textit{Tool-based visual reasoning}\hfill\textit{Inference: $45\times$}}} \\[0.3ex]
    Thyme
    & \cmark & \cmark & 1424
    & 66.6 & 63.1 & 71.2
    & 32.3 & 60.5 & 47.5 & \textbf{54.6} & 64.0 & 44.4
    & 75.3 & 58.0 \\
    
    DeepEyes
    & \xmark & \cmark & 1536
    & 72.8 & 69.2 & \textbf{88.5}
    & 28.5 & 39.5 & 47.5 & 43.2 & 57.3 & \underline{52.6}
    & \underline{91.8} & 59.1 \\
    
    \addlinespace[0.8ex]
    \midrule
    
    \multicolumn{15}{@{}c@{}}{\makebox[\zebratw][s]{\textit{Controlled baselines and ours}\hfill\textit{Inference: $1\times$}}} \\[0.3ex]
    
    Qwen2.5-VL-7B
    & \xmark & \xmark & --
    & 68.9 & 64.1 & 75.9
    & 24.6 & 56.4 & 45.9 & 47.2 & 64.7 & 43.6
    & 86.5 & 57.8 \\
    
    \rowcolor{gray!15}
    \textbf{V2T (Ours)}
    & \cmark
    & \xmark
    & 20
    & 69.3
    & 64.5
    & 80.1
    & 28.5
    & \underline{66.3}
    & 51.6
    & 47.7
    & \underline{66.0}
    & \underline{52.6}
    & \underline{91.8}
    & \underline{61.8} \\
    
    \rowcolor{gray!15}
    \textbf{V2T + GRPO (Ours)}
    & \cmark
    & \cmark
    & 80
    & \underline{73.5}
    & \textbf{71.3}
    & \underline{82.2}
    & \textbf{36.9}
    & 59.9
    & 45.1
    & \textbf{54.6}
    & 62.7
    & \textbf{55.6}
    & \textbf{92.0}
    & \textbf{63.4} \\
    
    \bottomrule
    \end{tabular}}%
    \immediate\write\csname @auxout\endcsname{\gdef\string\zebratw{\the\wd\zebratablebox}}%
    \begin{adjustbox}{max width=\textwidth,center}
    \usebox{\zebratablebox}
    \end{adjustbox}
    \vspace{-0.6em}
\end{table*}

%% file: experiment.tex
\section{Experiments}
\vspace{-1mm}

\paragraph{Training Datasets.}
We use the image–text interleaved CoT dataset \textbf{Zebra-CoT} \citep{li2025zebra}, a curated mixture spanning scientific diagrams, 2D visual puzzles, and visually grounded logic and strategic games. Following \citet{wang2025monet}, we apply a two-stage filtering process to remove trivial samples and noisy interleaved images. First, we use Qwen2.5-VL-7B with only the problem image and question, removing samples it answers correctly without interleaved images. Second, we use Qwen2.5-VL-72B with the interleaved images and remove samples it answers incorrectly, filtering out inaccurate images and overly difficult problems. This yields approximately 23K training samples, with 200 samples per category held out for testing.

\textbf{Implementation Details.}
For Zebra-CoT, we use Qwen2.5-VL-7B \citep{Qwen2.5-VL} as the base model and train both the teacher and the student with LoRA SFT \citep{hu2022lora}. The teacher uses the filtered Zebra-CoT set; the student uses its visual-search subset. We then run full-parameter GRPO \citep{shao2024deepseekmath} on TreeVGR-RL-37K \citep{wang2025traceable}, with $\lambda=0.1$, $\beta=0.5$, and $\eta=0.5$. SFT and GRPO hyperparameters are listed in Tables~\ref{tab:sft_hparams} and~\ref{tab:grpo_hparams} (Appendix~\ref{app:grpo_hparams}).

\textbf{Efficiency.}
  V2T requires approximately 20 NVIDIA H100 GPU-hours in total for teacher SFT and student distillation. The optional GRPO stage adds approximately 60 GPU-hours, bringing the total training cost to 80 GPU-hours. As shown in
  Table~\ref{tab:zebra_table2}, even V2T + GRPO uses $17.8\times$, $19.2\times$, and $42.4\times$ fewer GPU-hours than Thyme, DeepEyes, and Vision-R1-7B, respectively. At inference, V2T matches text-only reasoning at $1\times$, compared with $2\times$ for latent visual reasoning, $30\times$ for Bagel-Zebra and $45\times$ for tool-based methods.

\textbf{Baselines.} We compare V2T with text-based reasoning LVLMs (R1-VL-7B \citep{zhang2025r1}, VLAA-Thinker-7B \citep{chen2025sft}, VL-Rethinker-7B \citep{wang2025vl}, Vision-R1-7B \citep{huang2025vision}), latent visual reasoning (LVR-7B \citep{li2025latent}, MONET-7B \citep{wang2025monet}), the interleaved model Bagel-Zebra-7B \citep{li2025zebra}, and tool-based methods (Thyme \citep{zhang2025thyme}, DeepEyes \citep{zheng2025deepeyes}). The controlled baseline trained with our data in Table~\ref{tab:zebra_table2} is Qwen2.5-VL-7B. Text-Only SFT and Interleaved SFT, the teacher, are reported in Table~\ref{tab:attention_rl_ablation}. We also report V2T + GRPO. Details are provided in Appendix~\ref{app:experimental_details}.

\textbf{Evaluation Datasets.}
We evaluate on two complementary suites: a held-out Zebra-CoT set covering Graph Algorithm, RPM, Tetris, and Visual Search, and a broader visual evaluation suite including HR-Bench 4K/8K \citep{wang2024hrbench}, V*Bench \citep{wu2024vstarbench}, BLINK \citep{fu2024blink}, and POPE \citep{li2023pope}. We use VLMEvalKit \citep{duan2024vlmevalkit} with Qwen3-235B-A22B-Instruct-2507-FP8 \citep{yang2025qwen3} as the judge model. Evaluation details are provided in Appendix~\ref{app:experimental_details}.
We additionally conduct a judge-consistency check on selected benchmarks in Appendix~\ref{app:judge_consistency}. We set the maximum generation length to 2048 tokens.

\subsection{Main Results}
\textbf{Zebra-CoT held-out tasks.}
Table~\ref{tab:zebra_cot} reports results on the Zebra-CoT held-out subset, covering Graph Algorithm, RPM, Tetris, and Visual Search. V2T achieves the highest average accuracy of 75.5\%, outperforming the Interleaved SFT teacher by 14.3\% and the strongest existing baseline, VLAA-Thinker-7B, by 14.6\%. V2T performs best on all four held-out task categories, confirming the effectiveness of our visual distillation into text space.

\textbf{Broader visual evaluation.}
Table~\ref{tab:zebra_table2} reports results across high-resolution perception (HR-Bench 4K/8K), visual search (V*Bench), fine-grained perception (BLINK), and hallucination (POPE). Without GRPO, V2T achieves an average score of 61.8\%, outperforming the strongest existing visual reasoning models, Vision-R1-7B and DeepEyes, by 2.7\%. V2T also exceeds Qwen2.5-VL-7B by 4.0\%. Compared with Qwen2.5-VL-7B, V2T improves on every reported task, with the largest gains on BLINK Visual Correspondence ($+9.9\%$), Multi-view Reasoning ($+9.0\%$), Object Localization ($+5.7\%$), and POPE ($+5.3\%$). Comparisons with Text-Only SFT and Interleaved SFT are in Table~\ref{tab:attention_rl_ablation}. The results confirm that V2T distillation improves both visual grounding and overall generalization without requiring intermediate visual states at inference.

%% file: ablation.tex
\subsection{Ablation Studies}
\label{sec:ablation}
\vspace{-1mm}
\begin{table*}[t]
    \centering
    \caption{Ablation of attention supervision: attention distillation during SFT and the visual grounding reward during GRPO, compared with Text-Only SFT and Interleaved SFT, the teacher. Best and second-best results are highlighted in \textbf{bold} and \underline{underline}, respectively. Evaluation categories follow Table~\ref{tab:zebra_table2}.}
    \label{tab:attention_rl_ablation}
    \small
    \setlength{\tabcolsep}{0.8pt}
    \renewcommand{\arraystretch}{1.08}
    \begin{adjustbox}{max width=\textwidth,center}
    \begin{tabular}{@{}lccc cccccc cc@{}}
    \toprule
    \textbf{Method}
    & \textbf{HR4K}
    & \textbf{HR8K}
    & \textbf{V*}
    & \multicolumn{6}{c}{\textbf{BLINK}}
    & \textbf{POPE}
    & \textbf{Avg} \\
    \cmidrule(lr){5-10}
    & & &
    & Func. & Visual & Obj. Loc. & Foren. & Jig. & Multi.
    & & \\
    \midrule
    Text-Only SFT
    & 60.1 & 47.3 & 66.0
    & 25.4 & 60.5 & 50.0 & 52.3 & 63.3 & 45.1
    & 91.7 & 56.2 \\

    Interleaved SFT
    & 56.8 & 47.1 & 62.8
    & 26.9 & \underline{61.6} & 48.4 & 44.7 & 60.0 & 48.9
    & 91.6 & 54.9 \\

    \midrule
    \rowcolor{gray!15}
    \textbf{V2T (Ours)}
    & 69.3 & 64.5 & 80.1
    & 28.5 & \textbf{66.3} & \textbf{51.6} & 47.7 & \textbf{66.0} & \underline{52.6}
    & \underline{91.8} & 61.8 \\

    V2T w/o Attention Distillation
    & 59.0 & 49.6 & 63.4
    & \underline{31.5} & 60.5 & 46.7 & 48.5 & \underline{64.7} & \textbf{55.6}
    & 91.5 & 57.1 \\

    \midrule
    \rowcolor{gray!15}
    \textbf{V2T + GRPO (Ours)}
    & \textbf{73.5} & \underline{71.3} & \underline{82.2}
    & \textbf{36.9} & 59.9 & 45.1 & \underline{54.6} & 62.7 & \textbf{55.6}
    & \textbf{92.0} & \textbf{63.4} \\

    V2T + GRPO w/o Grounding Reward
    & \underline{73.1} & \textbf{71.4} & \textbf{84.3}
    & 20.8 & 55.2 & \underline{50.8} & \textbf{59.1} & 58.0 & \textbf{55.6}
    & \textbf{92.0} & \underline{62.0} \\
    \bottomrule
    \end{tabular}
    \end{adjustbox}
    \vspace{-0.6em}
\end{table*}

\textbf{Attention distillation and visual grounding reward.}
As shown in Table~\ref{tab:attention_rl_ablation}, V2T without attention distillation averages 57.1, above Text-Only SFT (56.2) and Interleaved SFT (54.9). Adding attention distillation improves the average from 57.1\% to 61.8\%, a gain of 4.7\%. The largest improvements occur on HR-Bench 4K ($+10.3\%$), HR-Bench 8K ($+14.9\%$), and V*Bench ($+16.7\%$), indicating that attention distillation is particularly important for high-resolution perception and visual search. During GRPO, the visual grounding reward increases the average from 62.0\% to 63.4\%. The improvement is especially pronounced on BLINK Functional Correspondence ($+16.1\%$), while performance varies across the other BLINK categories, suggesting that attention grounding and task-reward optimization provide complementary but task-dependent signals.

\textbf{Hyperparameter Ablations.}
We study the sensitivity to the logit distillation weight $\lambda$,
attention supervision weight $\beta$, and visual grounding reward weight $\eta$.
We use $\lambda=0.1$, $\beta=0.5$, and $\eta=0.5$ in our final configuration.
Full results are provided in Appendix~\ref{app:hparam-ablation}.

%% file: analysis.tex
\section{Error Analysis}
\label{sec:analysis}
Beyond aggregate accuracy, we analyze whether V2T genuinely internalizes visual grounding and which types of errors it reduces. We study this from two perspectives: attention alignment with ground-truth visual regions on test set, and error distributions across different failure categories. 
In this section we analyze V2T without the subsequent GRPO training. 

\textbf{Attention Analysis.} 
We compare attention patterns across V2T, the interleaved SFT model evaluated with and without reasoning images, and the text-only model. Following prior analyses showing that cross-modal visual information is primarily processed in the middle Transformer layers \citep{kaduri2025image,jiang2025devils}, we average attention over the middle 8 layers to obtain a more stable estimate of visual grounding. We take the maximum across heads and measure attention IoU between the top-$k$ attended pixels and the ground-truth bounding box, where $k$ matches the bounding-box area. Since attention to visual tokens is strongest in the first few generated tokens and rapidly decreases afterward \citep{ramazzina2025reasoning}, we focus on the first text token following each reasoning-image insertion point, consistent with the attention distillation in Section~\ref{sec:attention_distillation}. We evaluate on 100 random examples from Zebra-Visual Search. %

\begin{table}[h]
\centering
\caption{Attention IoU analysis. (Inf) and (Tra) denote the interleaved model evaluated without and with reasoning images respectively. %
V2T pays highest attention to the bounding boxes.}
\label{tab:attention_iou}
\vspace{-2mm}
\begin{small}
\begin{adjustbox}{max width=\linewidth,center}
\begin{tabular}{lccccc}
\toprule
\textbf{Model} & \textbf{Base Model} & \textbf{Text-Only SFT} & \textbf{Interleaved (Inf)} & \textbf{Interleaved (Tra)} & \textbf{V2T (Ours)} \\
\midrule
\textbf{IoU} & 0.191 & 0.215 & 0.223 & 0.168 %
& \textbf{0.269} \\
\bottomrule
\end{tabular}
\end{adjustbox}
\end{small}
\end{table}

\begin{figure*}[t]
\centering

\begin{minipage}[t]{0.3\textwidth}
    \centering
    \includegraphics[width=\linewidth,trim=0 0 0 220,clip]{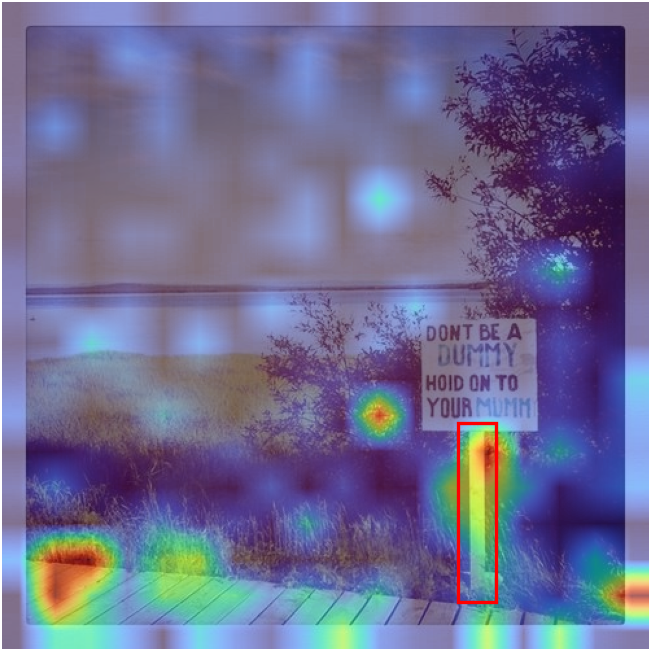}
    \caption*{(a) Text-Only SFT}
    \label{fig:attn_text_only}
\end{minipage}
\hfill
\begin{minipage}[t]{0.3\textwidth}
    \centering
    \includegraphics[width=\linewidth,trim=0 0 0 220,clip]{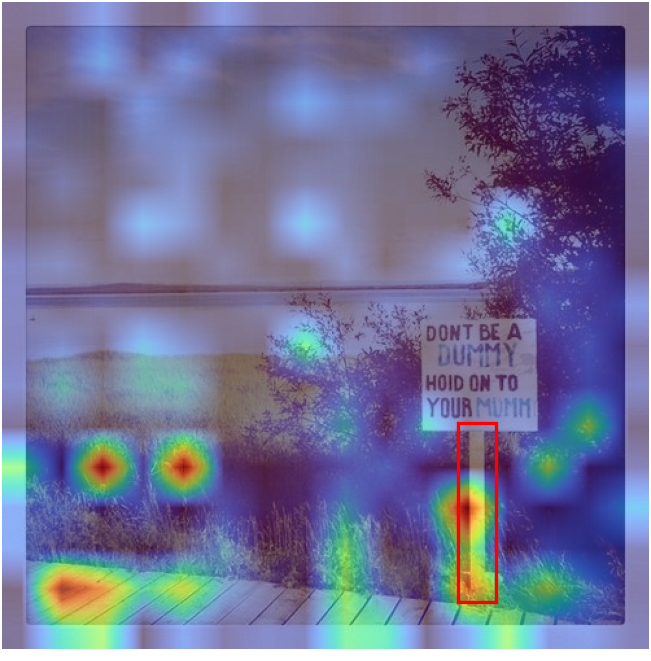}
    \caption*{(b) Interleaved SFT (teacher)}
    \label{fig:attn_interleaved}
\end{minipage}
\hfill
\begin{minipage}[t]{0.3\textwidth}
    \centering
    \includegraphics[width=\linewidth,trim=0 0 0 220,clip]{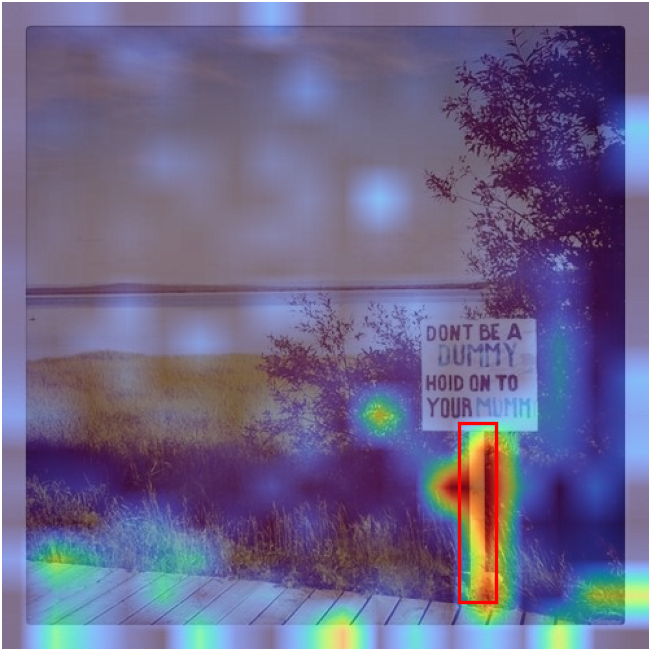}
    \caption*{(c) V2T (student)}
    \label{fig:attn_v2t}
\end{minipage}
\vspace{-2mm}
\caption{Qualitative comparison of attention maps among Text-Only SFT, Interleaved SFT, and V2T on the Visual Search split of Zebra-CoT. V2T attends more strongly to task-relevant visual regions, suggesting improved visual grounding compared to text-only and Interleaved SFT (teacher) models.}
\label{fig:attention_map_comparison}
\vspace{-4mm}
\end{figure*}

As shown in Table~\ref{tab:attention_iou} and Fig.~\ref{fig:attention_map_comparison}, Text-Only SFT only slightly improves over the base model ($0.215$ vs.\ $0.191$), suggesting that textual CoT supervision alone provides limited visual grounding. The interleaved model evaluated without reasoning images (Inf) reaches $0.223$, but when reasoning images are provided (Tra), its attention shifts to the reasoning image (IoU $0.234$) and its problem-image IoU drops to $0.168$, below the base model, indicating that the reasoning image can act as a shortcut. In contrast, V2T achieves the highest problem-image IoU ($0.269$), showing that V2T transfers genuine visual grounding into text space rather than collapsing into linguistic pattern matching.

\textbf{Error Analysis. }
To understand where V2T improves, we use GPT-4o to classify incorrect predictions into four categories: (1)~\textit{Perception and Grounding}, (2)~\textit{Spatial and Global Context}, (3)~\textit{Reasoning and Knowledge}, and (4)~\textit{Task and Answer Consistency}.
\vspace{-2mm}
\begin{table}[h]
\centering
\caption{Error distribution. V2T has the largest relative reduction in Spatial \& Global Context errors.}
\label{tab:error_analysis}
\vspace{-3mm}
\begin{small}
\begin{tabular}{lccc}
\toprule
\textbf{Category} & \textbf{Text-Only SFT} & \textbf{Interleaved SFT} & \textbf{V2T (Ours)} \\
\midrule
Perception \& Grounding & 366 & 382 & \textbf{364} \\
Spatial \& Global Context & 241 & 230 & \textbf{193} \\
Reasoning \& Knowledge & 941 & 916 & \textbf{872} \\
Task \& Answer Consistency & 199 & 209 & \textbf{179} \\
\midrule
\textbf{Total} & 1747 & 1737 & \textbf{1608} \\
\bottomrule
\end{tabular}
\end{small}
\end{table}

As shown in Table~\ref{tab:error_analysis}, V2T reduces total errors by $7.4\%$ relative to Interleaved SFT, with the largest reduction in Spatial and Global Context errors ($-16.1\%$) and Task and Answer Consistency errors ($-14.4\%$). This is consistent with the attention analysis: stronger grounding to the original problem image particularly improves scene-level spatial reasoning. Perception and Grounding errors remain similar across methods, while Reasoning and Knowledge errors still dominate, suggesting that V2T primarily improves visually grounded reasoning rather than basic perception or domain knowledge.

%% file: conclusion.tex
\vspace{-2mm}
\section{Conclusion}
\vspace{-2mm}
We propose V2T, a training framework that uses interleaved visual information as training-time supervision and distills its visual grounding into text-based chain-of-thought reasoning. On held-out and broader visual benchmarks, V2T outperforms text-only reasoning models, latent visual reasoning approaches, and interleaved models, including its teacher. These results show that visual reasoning can be internalized through distillation: at inference, V2T reasons entirely in text at the same $1\times$ cost as text-only models, without intermediate visual representations or their error propagation.
\looseness=-1

%% file: appendix.tex
\providecommand{\FloatBarrier}{\clearpage}

\section*{Impact Statement}

This paper presents work whose goal is to advance the field of machine learning. There are many potential societal consequences of our work, none that we feel must be specifically highlighted here.

\textbf{Limitations.}
While V2T achieves strong gains, several limitations remain. First, it requires interleaved multimodal reasoning data with paired intermediate images, which is expensive to collect and limited in scale. Second, the student inherits any biases or errors in the teacher's visually-grounded reasoning. Third, our error analysis shows that text-space distillation primarily improves spatial reasoning, with smaller gains on perception-level and domain-knowledge errors, indicating it cannot substitute for stronger underlying visual perception. Finally, due to computational constraints, all experiments use a single training run; multi-seed evaluation and larger model scales are left to future work.

\section{Experimental Details}

\subsection{SFT Hyperparameters}
We list the SFT hyperparameters in Table~\ref{tab:sft_hparams}. We use LoRA-based fine-tuning with rank $r=32$ and $\alpha=64$ on four NVIDIA H100 80GB GPUs. The overall batch size is 64 with four gradient-accumulation steps. To avoid out-of-memory errors on samples containing many interleaved images, we constrain each training sequence to between $512\times28\times28$ and $1280\times28\times28$ image pixels.

\begin{table}[H]
\centering
\caption{Hyperparameters for SFT.}
\label{tab:sft_hparams}
\small
\begin{tabular}{@{}l r@{}}
\toprule
\textbf{Hyperparameter} & \textbf{Value} \\
\midrule
LoRA rank & 32 \\
LoRA alpha & 64 \\
LoRA target modules & All \\
Learning rate & $1.0\times10^{-4}$ \\
Per-device batch size & 4 \\
Gradient accumulation steps & 4 \\
Training epochs & 2 \\
Learning-rate scheduler & Cosine \\
Warmup ratio & 0.1 \\
Maximum image pixels & $1280 \times 28 \times 28$ \\
Minimum image pixels & $512 \times 28 \times 28$ \\
Distillation coefficient $\lambda$ & 0.1 \\
Attention supervision coefficient $\beta$ & 0.5 \\
\bottomrule
\end{tabular}
\end{table}

\subsection{GRPO Hyperparameters}
\label{app:grpo_hparams}
The optional GRPO stage starts from the distilled V2T student and performs full-parameter training on TreeVGR-RL-37K \citep{wang2025traceable}. For each prompt, we sample five rollouts. The reward is $r = r_{\mathrm{task}} + \eta r_{\mathrm{ground}}$ (Section~\ref{sec:attention_grounding_reward}), where $r_{\mathrm{task}}$ combines a binary answer-correctness reward with a format reward weighted by $0.1$, and $\eta=0.5$. We train for 160 steps with EasyR1 on four NVIDIA H100 GPUs, which takes approximately 60 GPU-hours. Table~\ref{tab:grpo_hparams} summarizes the settings.

\begin{table}[H]
\centering
\caption{Hyperparameters for GRPO.}
\label{tab:grpo_hparams}
\small
\begin{tabular}{@{}l r@{}}
\toprule
\textbf{Hyperparameter} & \textbf{Value} \\
\midrule
Initialization & V2T student \\
Training data & TreeVGR-RL-37K \\
Fine-tuning & Full-parameter \\
Learning rate & $1.0\times10^{-6}$ \\
Learning-rate scheduler & Constant \\
KL coefficient (low-variance KL) & 0.01 \\
Rollout batch size (prompts) & 512 \\
Global batch size & 128 \\
PPO epochs & 1 \\
Training steps & 160 \\
Rollouts per prompt & 5 \\
Rollout temperature & 1.0 \\
Maximum prompt / response length & 2048 / 2048 \\
Format reward weight & 0.1 \\
Visual grounding reward weight $\eta$ & 0.5 \\
\bottomrule
\end{tabular}
\end{table}

\subsection{Evaluation and Baseline Details}
\label{app:experimental_details}

\paragraph{Evaluation Datasets.}
We use two complementary evaluation suites. First, we evaluate on a held-out subset of Zebra-CoT comprising 200 randomly sampled examples from each of Graph Algorithm, RPM, Tetris, and Visual Search (800 in total). Second, we evaluate broader visual generalization on HR-Bench 4K and 8K \citep{wang2024hrbench}, V*Bench \citep{wu2024vstarbench}, BLINK \citep{fu2024blink}, and POPE \citep{li2023pope}. HR-Bench evaluates fine-grained perception at high image resolutions, while V*Bench measures visual search and spatial grounding. From BLINK, we report Functional Correspondence, Visual Correspondence, Object Localization, Forensic Detection, Jigsaw, and Multi-view Reasoning. POPE evaluates object hallucination. Together, these benchmarks cover in-domain multimodal reasoning as well as out-of-domain high-resolution perception, visual correspondence, localization, compositional and multi-view reasoning, and hallucination.

We adopt VLMEvalKit \citep{duan2024vlmevalkit} for evaluation and use Qwen3-235B-A22B-Instruct-2507-FP8 \citep{yang2025qwen3} as the judge model.

\paragraph{Baselines.}
All baselines are 7B-scale models. We group them by how they incorporate visual information during reasoning. (1)~\emph{Text-only visual reasoning}: R1-VL-7B \citep{zhang2025r1}, VLAA-Thinker-7B \citep{chen2025sft}, VL-Rethinker-7B \citep{wang2025vl}, and Vision-R1-7B \citep{huang2025vision}, which generate textual CoTs over the original image after SFT and/or RL post-training. (2)~\emph{Latent visual reasoning}: LVR-7B \citep{li2025latent} and MONET-7B \citep{wang2025monet}, which interleave learned latent visual tokens with textual reasoning. (3)~\emph{Unified interleaved generation}: Bagel-Zebra-7B \citep{li2025zebra}, which generates explicit intermediate images and is trained on the full Zebra-CoT dataset; we therefore exclude it from the Zebra-CoT held-out evaluation. (4)~\emph{Tool-based visual reasoning}: Thyme \citep{zhang2025thyme} and DeepEyes \citep{zheng2025deepeyes}, which operate on the image through tool calls (e.g., cropping) during reasoning. To match our single-pass inference setting, we reduce Thyme's default five answering attempts to a single attempt. Qwen2.5-VL-7B \citep{Qwen2.5-VL} serves as the controlled base model for Text-Only SFT, Interleaved SFT (the teacher), and V2T. For all baselines, we use the official checkpoints and the VLMEvalKit evaluation pipeline. Bagel-Zebra-7B is served with vLLM, which supports only its image-understanding pathway, so it is evaluated with text-only reasoning. Note that R1-VL-7B is built on Qwen2-VL rather than Qwen2.5-VL.

\paragraph{Training and inference cost.}
Training costs in Table~\ref{tab:zebra_table2} are taken from the GPU-hours reported by each method; entries marked with $^{*}$ are estimated from the reported training configurations and public implementations. Tool-based methods also use more expensive RL: for example, DeepEyes (1536 GPU-hours) samples 16 rollouts per prompt with repeated image cropping and re-encoding, compared with five text-only rollouts in our GRPO stage. Thyme and DeepEyes also train on V* data, which shares its distribution with V*Bench, whereas V2T is trained on Zebra-CoT without comparable V* examples. Inference cost is normalized to standard text-only autoregressive reasoning.

\Needspace{0.40\textheight}
\subsection{Filtered Zebra-CoT Composition}
The filtered training set contains 23,260 examples. Table~\ref{tab:zebra-distribution} reports its composition by source subset.

\begin{table}[H]
\centering
\caption{Composition of the filtered Zebra-CoT training set.}
\label{tab:zebra-distribution}
\small
\begin{tabular}{@{}l r@{}}
\toprule
\textbf{Subset} & \textbf{Percentage} \\
\midrule
Visual logic and strategic games -- ARC-AGI-New & 0.056\% \\
Scientific reasoning -- Geometry & 1.045\% \\
Scientific reasoning -- Competitive Programming & 1.040\% \\
Visual logic and strategic games -- RPM & 2.266\% \\
Visual logic and strategic games -- ARC-AGI & 0.056\% \\
3D visual reasoning -- Embodied CoT & 0.172\% \\
3D visual reasoning -- Multi-hop Object Counting & 5.864\% \\
Scientific reasoning -- Chemistry & 12.455\% \\
2D visual reasoning -- Visual Search & 33.780\% \\
Visual logic and strategic games -- Maze & 16.062\% \\
Scientific reasoning -- Physics & 0.052\% \\
3D visual reasoning -- Robot Planning & 0.254\% \\
Visual logic and strategic games -- Tetris & 2.171\% \\
Scientific reasoning -- Graph Algorithms & 8.060\% \\
Visual logic and strategic games -- Ciphers & 0.080\% \\
Visual logic and strategic games -- Chess & 15.900\% \\
Visual logic and strategic games -- Connect Four & 0.688\% \\
\bottomrule
\end{tabular}
\end{table}

\subsection{Hyperparameter Ablations}
\label{app:hparam-ablation}

We study the effect of three loss weights: the logit distillation weight
$\lambda$, the attention supervision weight $\beta$, and the visual grounding
reward weight $\eta$.

\begin{figure}[!htbp]
    \centering
    \includegraphics[width=0.72\linewidth]{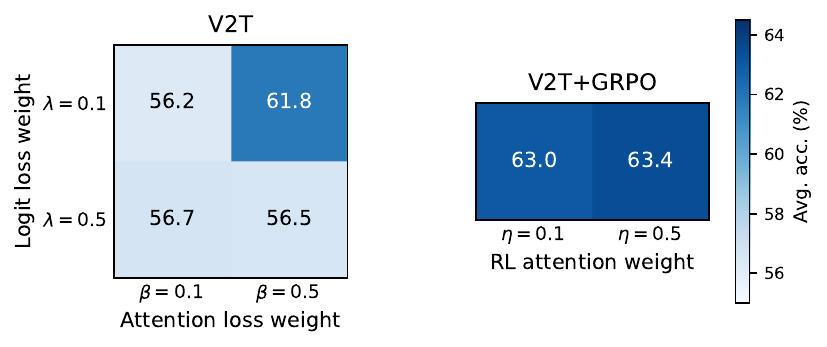}
    \caption{
    Hyperparameter ablations for V2T and V2T+GRPO.
    Left: average accuracy of V2T when varying the logit distillation
    weight $\lambda$ and the attention supervision weight $\beta$.
    Right: average accuracy of V2T+GRPO when varying the visual grounding
    reward weight $\eta$.
    }
    \label{fig:hparam-ablation}
\end{figure}

For V2T, we jointly vary $\lambda$ and $\beta$ over $\{0.1,0.5\}$.
As shown in Figure~\ref{fig:hparam-ablation}, the best performance is
obtained with $\lambda=0.1$ and $\beta=0.5$, achieving an average accuracy
of 61.8\%. The other configurations, $(\lambda,\beta)=(0.1,0.1)$,
$(0.5,0.1)$, and $(0.5,0.5)$, achieve 56.2\%, 56.7\%, and 56.5\%,
respectively.

For V2T+GRPO, we further vary the visual grounding reward weight $\eta$
over $\{0.1,0.5\}$. $\eta=0.5$ achieves the best average accuracy of
63.4\%, compared with 63.0\% for $\eta=0.1$; we use $\eta=0.5$ in our
final configuration.

\section{Controlled Analysis of Interleaved Visual Reasoning}
\label{app:motivation_analysis}

The motivation for V2T is that intermediate visual states can provide useful supervision during training, while repeatedly conditioning the autoregressive textual reasoning process on additional visual representations can also introduce interference. We therefore separate two effects that are closely related but conceptually distinct. First, as more intermediate reasoning images are inserted into a chain of thought, the loss of subsequent textual reasoning can increase. Second, visual tokens need not be equally relevant to the current reasoning step: background regions and other unmatched visual content can further interfere with textual reasoning even when the useful visual content is unchanged. The latter effect is consistent with prior observations that imperfect cross-modal alignment remains a challenge for vision--language models \citep{tang2025seam,xu2026crossmathbench,huang2025deciphering}.

\paragraph{Modality Switching and Number of Interleaved Images.}
We first compare the per-token loss of textual reasoning under two conditions: (1) the textual CoT is conditioned on its interleaved reasoning images, and (2) the same textual CoT is evaluated without the interleaved reasoning images. We conduct this analysis on held-out Zebra-CoT examples from Checkers, Connect Four, and Object Counting. Across all three categories, the loss gap becomes substantially larger near the end of the reasoning trajectory. At 25\%, 50\%, 75\%, and 100\% of the CoT, the loss gaps are 0.02, 1.09, 2.68, and 4.53 for Checkers; 0.11, 0.48, 1.77, and 2.29 for Connect Four; and 0.10, 0.18, 3.02, and 3.91 for Object Counting. We observe the same trend when varying the number of interleaved reasoning images directly: the mean loss gap increases from $+0.38$ with three images to $+2.19$, $+5.15$, and $+5.99$ with five, seven, and nine images, respectively. These results provide direct empirical evidence that repeatedly introducing intermediate visual states can increasingly disrupt subsequent textual continuation even when the images contain reasoning-relevant information.

\paragraph{Irrelevant Visual Content.}
We next isolate a second source of interference: visual content that is not useful for the corresponding textual reasoning. For each reasoning image, we place the original image on a larger gray canvas so that the useful reasoning content is unchanged while the amount of irrelevant visual content is increased by 50\%. We then compare the textual-CoT loss against the same text-only condition used above. The loss gap increases for all three categories: from 4.53 to 5.12 for Checkers, from 3.91 to 9.90 for Object Counting, and from 2.29 to 3.50 for Connect Four. This shows that visually unmatched content can further interfere with textual reasoning even when the useful visual information is preserved.

\paragraph{Generation-Level Prefix Analysis.}
To verify that these effects are reflected in generation quality rather than only token-level loss, we conduct an additional experiment in which the model is given the first 3, 5, or 7 steps of a ground-truth multimodal reasoning trajectory and then generates the remaining textual reasoning and final answer. We compare three conditions: retaining the original reasoning images, removing the reasoning images while keeping exactly the same textual CoT context, and adding 50\% gray background to the original reasoning images. For Object Counting, the accuracies for With Image / No Image / +50\% Gray are 64/56/46 with three images, 60/62/56 with five images, and 60/66/58 with seven images. For Connect Four, the corresponding results are 8/4/4, 12/12/10, and 14/16/10.

The generation results reveal two complementary trends. When only a few reasoning images are present, the visual states can still provide useful information. However, as the number of interleaved images grows, retaining the same textual reasoning while removing the intermediate images matches or outperforms the interleaved condition. Moreover, increasing visually unmatched content consistently reduces accuracy. Together with the loss analysis, these experiments motivate V2T: intermediate images are valuable as training-time supervision, but their useful reasoning signal can be distilled into text space so that inference does not require repeated visual conditioning or carry the additional interference introduced by irrelevant visual tokens.

\section{Baseline Sanity Check}
\subsection{Math-Focused Reasoning Models}
\label{app:vlaa_math}

We additionally examine VLAA-Thinker-7B on the mathematical reasoning benchmarks emphasized by its training setup. As shown in Table~\ref{tab:vlaa_math_sanity}, VLAA-Thinker-7B consistently exceeds the Qwen2.5-VL-7B base model on MathVerse \citep{zhang2024mathverse}, MathVista \citep{lu2023mathvista}, MM-Math \citep{sun2024mmmath}, and DynaMath \citep{zou2025dynamath}. This confirms that the lower performance observed for VLAA-Thinker-7B on our broader visual-reasoning suite is not due to a generally weak checkpoint. Instead, the results suggest that its math-focused training improves mathematical reasoning while transferring less strongly to the more general visual tasks considered in our evaluation.

\begin{table}[H]
\centering
\caption{Performance of Qwen2.5-VL-7B and VLAA-Thinker-7B on math-focused reasoning benchmarks.}
\label{tab:vlaa_math_sanity}
\small
\begin{tabular}{lcccc}
\toprule
\textbf{Model} & \textbf{MathVerse} & \textbf{MathVista} & \textbf{MM-Math} & \textbf{DynaMath} \\
\midrule
Qwen2.5-VL-7B & 43.9 & 69.1 & 41.1 & 56.5 \\
VLAA-Thinker-7B & \textbf{45.8} & \textbf{71.4} & \textbf{43.1} & \textbf{57.3} \\
\bottomrule
\end{tabular}
\end{table}

\section{Judge Consistency Analysis}
\label{app:judge_consistency}

To verify that our evaluation is not sensitive to the choice of judge model, we additionally re-evaluate the same outputs on selected benchmarks using GPT-4o \citep{openai2024gpt4o} and compare the results with our Qwen3-235B-A22B-Instruct-2507-FP8 judge. Across HR-Bench 4K, HR-Bench 8K, and V*Bench, the two judges produce nearly identical scores, with a maximum difference of 0.5 percentage points.

\begin{table}[H]
\centering
\caption{Judge-consistency comparison on selected benchmarks. GPT-4o is used only for this cross-check; Qwen3-235B-A22B-Instruct-2507-FP8 remains the judge used in our main evaluation.}
\label{tab:judge_consistency}
\small
\setlength{\tabcolsep}{5pt}
\begin{tabular}{llccc}
\toprule
\textbf{Model} & \textbf{Judge} & \textbf{HR-Bench 4K} & \textbf{HR-Bench 8K} & \textbf{V*Bench} \\
\midrule
Qwen2.5-VL-7B & Qwen3-235B & 68.9 & 64.1 & 75.9 \\
Qwen2.5-VL-7B & GPT-4o & 68.9 & 64.0 & 75.9 \\
\midrule
V2T + GRPO & Qwen3-235B & 73.5 & 71.3 & 82.2 \\
V2T + GRPO & GPT-4o & 73.1 & 71.2 & 81.7 \\
\bottomrule
\end{tabular}
\end{table}
\noindent
These results show that the reported performance is stable across the two judge models on the evaluated subset.

\Needspace{0.62\textheight}
\section{Qualitative Comparisons}
\subsection{Traffic-Light Recognition}
\begin{figure}[H]
\centering
\begin{minipage}[t]{0.52\linewidth}
\vspace{0pt}
\includegraphics[width=\linewidth]{images/RWQA_375.jpg}
\end{minipage}\hfill
\begin{minipage}[t]{0.46\linewidth}
\vspace{0pt}
\begin{tcolorbox}[vctpanel,colback=black!2]
\RaggedRight
\RaggedRight
\textbf{Question:} What color is the nearest set of traffic lights? \\
\textbf{Options:} A. Red, B. Green, C. Yellow \\
\textbf{Answer (GT):} A \\
\textbf{Base Model Problem Category:} object\_\allowbreak presence\_\allowbreak type\_\allowbreak or\_\allowbreak motion\_\allowbreak misread \\
\textbf{Interleaved Model Problem Category:} object\_\allowbreak presence\_\allowbreak type\_\allowbreak or\_\allowbreak motion\_\allowbreak misread
\end{tcolorbox}
\begin{tcolorbox}[vctpanel,title={Base model generation}]
\RaggedRight
B. Green
\end{tcolorbox}
\end{minipage}
\par\smallskip
\begin{minipage}[t]{0.485\linewidth}
\vspace{0pt}
\begin{tcolorbox}[vctpanel,title={Interleaved SFT generation}]
\RaggedRight
THOUGHT 0: To determine the color of the traffic lights, I need to locate them within the provided image.\par
THOUGHT 1: I will highlight the relevant area containing the traffic lights to focus my analysis.\par
THOUGHT 2: The highlighted region clearly shows the traffic lights and their illuminated state.\par
THOUGHT 3: Observing the illuminated lights in the highlighted area reveals their color.\par

\medskip
The final answer is: B. Green
\end{tcolorbox}
\end{minipage}\hfill
\begin{minipage}[t]{0.485\linewidth}
\vspace{0pt}
\begin{tcolorbox}[vctpanel,title={V2T generation}]
\RaggedRight
THOUGHT 0: To determine the color of the traffic lights, I need to locate them within the provided image.\par
THOUGHT 1: The traffic lights are small and distant in the original image. To clearly identify their color, I will generate a zoomed-in view of this specific area.\par
THOUGHT 2: The zoomed-in image clearly shows the traffic lights. Upon closer inspection, they appear to be illuminated in red.\par
THOUGHT 3: Based on the visual evidence from the magnified section, the traffic lights are red.\par

\medskip
The final answer is: A. Red
\end{tcolorbox}
\end{minipage}

\caption{\textbf{Qualitative comparison on a traffic-light recognition example.}
The top panel shows the original question image. The base model predicts \textit{Green}. The interleaved SFT model also predicts \textit{Green} after performing step-wise reasoning over the scene. In contrast, V2T focuses on the distant traffic lights and correctly predicts \textit{Red}, matching the ground-truth answer. Phrases such as ``zoomed-in view'' are inherited from the teacher's interleaved CoT format; no intermediate image is generated or provided, and V2T reasons only over the original image.}
\label{fig:traffic_light_case_study}
\end{figure}

\clearpage
\subsection{Flower Counting}
\begin{figure}[H]
\centering
\begin{minipage}[t]{0.37\linewidth}
\vspace{0pt}
\includegraphics[width=\linewidth]{images/RWQA_523.jpg}
\end{minipage}\hfill
\begin{minipage}[t]{0.61\linewidth}
\vspace{0pt}
\begin{tcolorbox}[vctpanel,colback=black!2]
\RaggedRight
\textbf{Question:} How many white flowers are in this image?\\
\textbf{Options:}\\
A. There are no white flowers in this image.\\
B. There is 1 white flower in this image.\\
C. There are 2 white flowers in this image.\\
\textbf{Answer (GT):} C\\
\textbf{Base Model Problem Category:} \texttt{counting\_\allowbreak or\_\allowbreak quantity\_\allowbreak error}\\
\textbf{Interleaved Model Problem Category:} \texttt{counting\_\allowbreak or\_\allowbreak quantity\_\allowbreak error}
\end{tcolorbox}
\begin{tcolorbox}[vctpanel,title={Base model generation}]
\RaggedRight
B. There is 1 white flower in this image.
\end{tcolorbox}
\end{minipage}
\par\smallskip
\begin{minipage}[t]{0.485\linewidth}
\vspace{0pt}
\begin{tcolorbox}[vctpanel,title={Interleaved SFT generation}]
\RaggedRight
THOUGHT 0: I need to carefully examine the provided image to identify and count all visible white flowers.\par
THOUGHT 1: To accurately determine the number of white flowers, I will focus on the central floral arrangement in the image.\par
THOUGHT 2: Upon close inspection of the main flower cluster, it is clear that there is only one prominent white flower present.\par
THOUGHT 3: Based on this observation, the total count of white flowers in the image is one.\par

\medskip
\textbf{The final answer is:} B
\end{tcolorbox}
\end{minipage}\hfill
\begin{minipage}[t]{0.485\linewidth}
\vspace{0pt}
\begin{tcolorbox}[vctpanel,title={V2T generation}]
\RaggedRight
THOUGHT 0: I need to examine the provided image to identify all visible white flowers.\par
THOUGHT 1: To clearly discern the number of white flowers, I will focus on the central floral arrangement.\par
THOUGHT 2: The zoomed-in view reveals two distinct white flowers, both appearing to be chrysanthemums.\par
THOUGHT 3: Based on this clear observation, I can now determine the total count of white flowers.\par

\medskip
\textbf{The final answer is:} C. There are 2 white flowers in this image.
\end{tcolorbox}
\end{minipage}

\caption{\textbf{Qualitative comparison on a counting example.}
The question image contains two visible white flowers in the central arrangement. The base model and the interleaved SFT model both under-count the flowers and predict only one white flower. In contrast, V2T focuses on the central floral arrangement, identifies two distinct white flowers, and correctly predicts the ground-truth answer. Phrases such as ``zoomed-in view'' are inherited from the teacher's interleaved CoT format; no intermediate image is generated or provided, and V2T reasons only over the original image.}
\label{fig:white_flower_counting_case_study}
\end{figure}

\FloatBarrier
\section{Interleaved CoT Examples}
\begin{figure}[p]
\centering

\setlength{\fboxsep}{1pt}     %
\setlength{\fboxrule}{0.5pt}  %

\begin{adjustbox}{max width=0.95\linewidth, max totalheight=0.92\textheight, center}
\fcolorbox{black}{white}{%
  \begin{minipage}{0.97\linewidth}
    \setlength{\parindent}{0pt}
    \setlength{\parskip}{2pt}

    \colorbox{gray!25}{%
      \parbox{\dimexpr\linewidth-2\fboxsep\relax}{\centering\tiny\textbf{
        Zebra-CoT Example 1:
      }}%
    }

    \vspace{4pt}

    \tiny
    \raggedright

    \textbf{Question:} What is the main organic product when
    \texttt{ethyl prop-2-enoate (C=CC(=O)OCC)} and
    \texttt{4-bromo-2-chloro-1-(trifluoromethyl)benzene (FC(F)(F)c1ccc(Br)cc1Cl)}
    are combined?

    \vspace{5pt}
    \raggedright
    \textbf{Image-text Interleaved CoT:}

    \justifying
    \textbf{THOUGHT 0:} The problem asks for the main organic product from the reaction of
    ethyl prop-2-enoate (an alkene derivative) and 4-bromo-2-chloro-1-(trifluoromethyl)benzene
    (an aryl halide). This combination suggests a palladium-catalyzed Heck reaction.

    \textbf{THOUGHT 1:} To understand the reaction, first visualize the structure of ethyl
    prop-2-enoate based on its SMILES \texttt{C=CC(=O)OCC}.

    \vspace{3pt}
    \centering
    \includegraphics[width=0.34\linewidth]{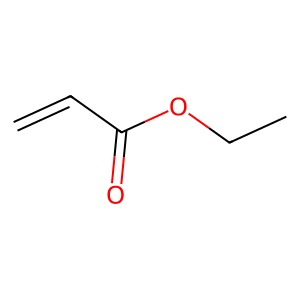}

    \vspace{4pt}
    \raggedright
    \justifying
    \textbf{THOUGHT 2:} The image shows ethyl prop-2-enoate, an $\alpha,\beta$-unsaturated ester
    (ethyl acrylate). The key reactive site is the terminal C=C bond.

    \textbf{THOUGHT 3:} Next, visualize the structure of
    4-bromo-2-chloro-1-(trifluoromethyl)benzene from \texttt{FC(F)(F)c1ccc(Br)cc1Cl}.

    \vspace{3pt}
    \centering
    \includegraphics[width=0.34\linewidth]{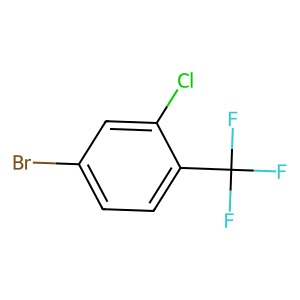}

    \vspace{4pt}
    \raggedright
    \justifying
    \textbf{THOUGHT 4:} The second reactant is an aryl halide containing both Br and Cl, plus a
    trifluoromethyl group.

    \textbf{THOUGHT 5:} In a Heck coupling, oxidative addition typically occurs at C--Br rather
    than C--Cl. Thus, coupling happens at the bromine position. The next image shows both reactants.

    \vspace{3pt}
    \centering
    \includegraphics[width=0.34\linewidth]{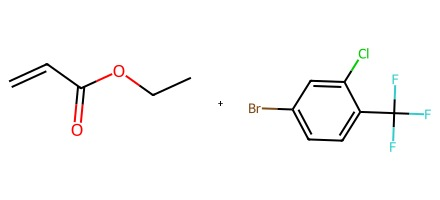}

    \vspace{4pt}
    \raggedright
    \justifying
    \textbf{THOUGHT 6:} The aryl group (2-chloro-4-(trifluoromethyl)phenyl) couples to ethyl
    acrylate. Regioselectivity places the aryl at the terminal carbon of the alkene.

    \textbf{THOUGHT 7:} After $\beta$-hydride elimination, the product is the substituted alkene,
    typically with trans (E) stereochemistry. Visualize the final product.

    \vspace{3pt}
    \centering
    \includegraphics[width=0.34\linewidth]{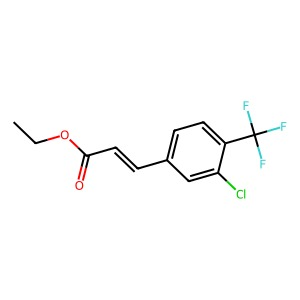}

    \vspace{4pt}
    \raggedright
    \justifying
    \textbf{THOUGHT 8:} The final image depicts the coupled product with a new C=C bond in the
    acrylate side chain, typically (E). The final answer is:
    \textbf{ethyl (E)-3-(2-chloro-4-(trifluoromethyl)phenyl)prop-2-enoate}.

  \end{minipage}%
}
\end{adjustbox}

\vspace{-6pt}
\caption{Generating intermediate visual states and interleaving grounded reasoning for chemistry.}
\label{fig:zebra_case_example2}
\end{figure}

\begin{figure}[p]
\centering

\setlength{\fboxsep}{1pt}     %
\setlength{\fboxrule}{0.5pt}  %

\begin{adjustbox}{max width=0.95\linewidth, max totalheight=0.92\textheight, center}
\fcolorbox{black}{white}{%
  \begin{minipage}{0.98\linewidth}
    \setlength{\parindent}{0pt}
    \setlength{\parskip}{2pt}

    \colorbox{gray!25}{%
      \parbox{\dimexpr\linewidth-2\fboxsep\relax}{\centering\tiny\textbf{
        Zebra-CoT Example 2:
      }}%
    }

    \vspace{4pt}

    \tiny
    \raggedright

    \textbf{Question:} What is the best move for Black to play?
    
    A: Bf5+
    B: h6
    C: Bb4
    D: Be6

    \vspace{3pt}
    \textbf{Question Image:}

    \vspace{2pt}
    \centering
    \includegraphics[width=0.25\linewidth]{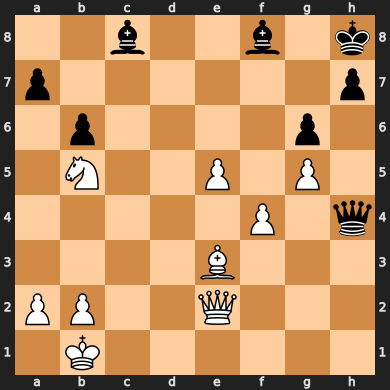}

    \vspace{5pt}
    \raggedright
    \textbf{Image-text Interleaved CoT:}

    \justifying
    \textbf{THOUGHT 0:} Analyze the initial position shown in the diagram. Black needs to find the best move among the given options. 

    \textbf{THOUGHT 1:} Consider option A, Bf5+. Visualize the position after this move.

    \vspace{3pt}
    \centering
    \includegraphics[width=0.25\linewidth]{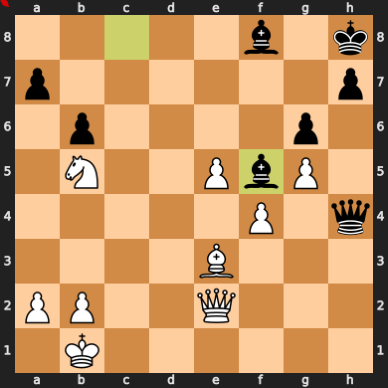}

    \vspace{4pt}
    \raggedright
    \justifying
    \textbf{THOUGHT 2:} The diagram shows the position after Black plays Bf5. 

    \textbf{THOUGHT 3:} Consider option B, h6. Visualize the position after this move.

    \vspace{3pt}
    \centering
    \includegraphics[width=0.25\linewidth]{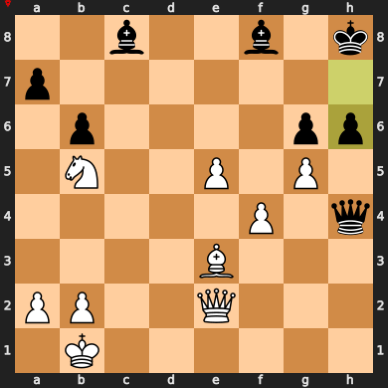}

    \vspace{4pt}
    \raggedright
    \justifying
    \textbf{THOUGHT 4:} The diagram shows the position after Black plays h6.

    \textbf{THOUGHT 5:} Consider option C, Bb4. Visualize the position after this move.

    \vspace{3pt}
    \centering
    \includegraphics[width=0.25\linewidth]{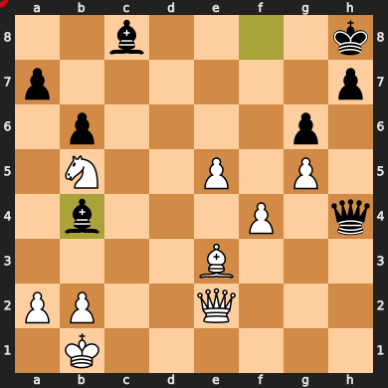}

    \vspace{4pt}
    \raggedright
    \justifying
    \textbf{THOUGHT 6:} The diagram shows the position after Black plays Bb4.

    \textbf{THOUGHT 7:} Consider option D, Be6. Visualize the position after this move.

    \vspace{3pt}
    \centering
    \includegraphics[width=0.25\linewidth]{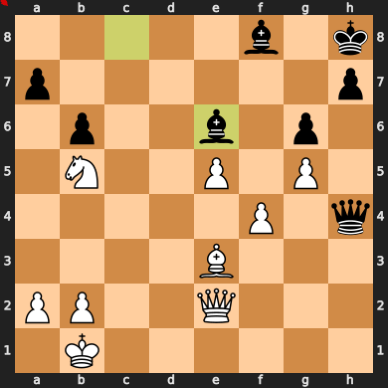}

    \vspace{4pt}
    \raggedright
    \justifying
    \textbf{THOUGHT 8:} The diagram shows the position after Black plays Be6.
    
    \textbf{THOUGHT 9:} Based on the evaluation of these options, the best move for Black is Bf5+ (Option A).
    
    The final answer is: A

  \end{minipage}%
}
\end{adjustbox}

\vspace{-6pt}
\caption{Generating intermediate visual states and interleaving grounded reasoning for chess.}
\label{fig:zebra_case_example_chess}
\end{figure}

Figures~\ref{fig:zebra_case_example2} and~\ref{fig:zebra_case_example_chess} show complete image--text interleaved reasoning examples from the chemistry and chess subsets of Zebra-CoT.

\clearpage

\clearpage